\pdfoutput=1

\documentclass[sigconf]{acmart}
\AtBeginDocument{%
  \providecommand\BibTeX{{%
    \normalfont B\kern-0.5em{\scshape i\kern-0.25em b}\kern-0.8em\TeX}}}

\setcopyright{none}
\renewcommand\footnotetextcopyrightpermission[1]{}
\acmConference[CHI In2Writing Workshop '23]{CHI '23 Workshop on Intelligent and Interactive Writing Assistants}{April 23, 2023}{Hamburg, Germany}

\usepackage{caption}
\usepackage{subcaption}
\usepackage{booktabs}
\usepackage{xcolor}
\usepackage{colortbl}

\definecolor{revc}{RGB}{0, 0, 0} 
\newcommand{\rev}[1]{{\color{revc}#1}}

\newcommand{\system}{Towards Explainable AI Writing Assistants for Non-native English Speakers}

\begin{document}

\title[\system{}]{Towards Explainable AI Writing Assistants \\ for Non-native English Speakers} 

\author{Yewon Kim}
\affiliation{%
  \institution{KAIST}
  \streetaddress{}
  \city{}
  \country{Republic of Korea}}
\email{yewon.e.kim@kaist.ac.kr}

\author{Mina Lee}
\affiliation{%
  \institution{Stanford University}
  \city{}
  \country{United States}
}
\email{minalee@cs.stanford.edu}

\author{Donghwi Kim}
\affiliation{%
  \institution{KAIST}
  \city{}
  \country{Republic of Korea}
}
\email{dhkim09@kaist.ac.kr}

\author{Sung-Ju Lee}
\affiliation{%
  \institution{KAIST}
  \city{}
  \country{Republic of Korea}
}
\email{profsj@kaist.ac.kr}


\begin{abstract}

We highlight the challenges faced by non-native speakers when using AI writing assistants to paraphrase text. 
Through an interview study with 15 non-native English speakers (NNESs) \rev{with varying levels of English proficiency}, we observe that they face difficulties in assessing paraphrased texts generated by AI writing assistants, \rev{largely due to the lack of explanations accompanying the suggested paraphrases.} Furthermore, we examine their strategies to assess AI-generated texts \rev{in the absence of such explanations}. Drawing on the needs of NNESs identified in our interview, we propose four potential user interfaces to enhance the writing experience of NNESs using AI writing assistants. \rev{The proposed designs focus on incorporating explanations to better support NNESs in understanding and evaluating the AI-generated paraphrasing suggestions.}
\end{abstract}

\maketitle

\section{Introduction}
\label{section:introduction}

\begin{figure}[t]
    \centering
    \includegraphics[width=0.95\linewidth]{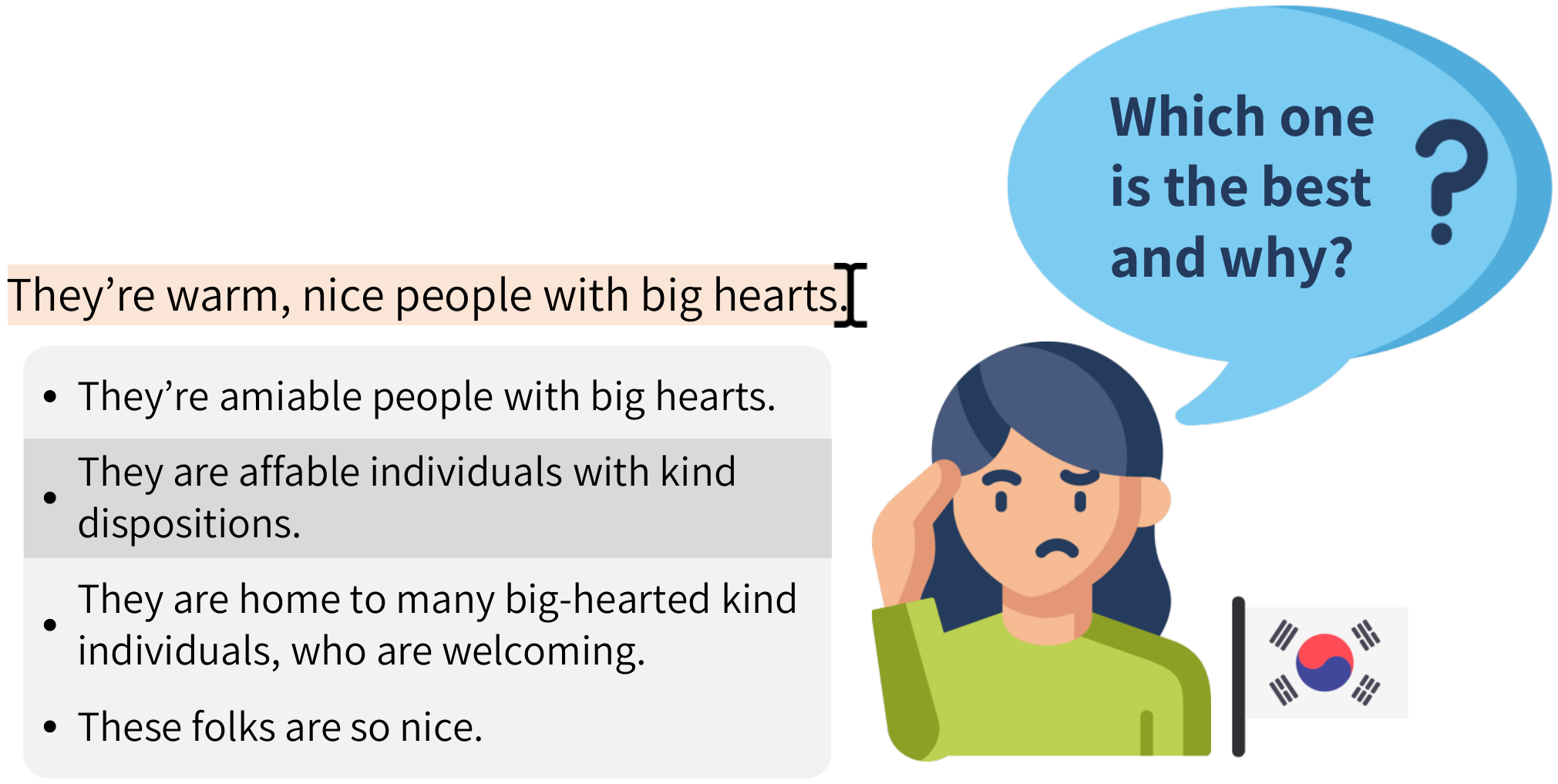}
    \vspace{-0.2cm}
    \caption{AI tools for paraphrasing often generate multiple suggestions given a user text as input~\cite{wordtune, quillbot}. In our interview, we observe that non-native English speakers (NNESs) struggle to assess and compare the quality of suggestions, particularly when suggestions lack explanations.
    }
    \label{fig:concept}
    \vspace{-.5cm}
\end{figure}

Effective written communication skill is an integral part of academic and professional success~\cite{doi:10.1080/03098260902734943, 10.1145/3173574.3173596}. However, non-native English speakers (NNESs) often struggle to achieve the desired level of linguistic accuracy and complexity when writing in English~\cite{fareed2016esl, nunan1989designing}. Despite their considerable efforts, \rev{they are prone to making errors and face difficulties in using diverse vocabularies}~\cite{polio1997measures, ortega2003syntactic}. Moreover, they have a hard time selecting appropriate words or phrases in various contexts due to different cultural backgrounds~\cite{varonis1985miscommunication, emailcommunication}. 
To improve their writing, many NNESs resort to AI writing assistants, such as grammar and style checkers~\cite{grammarly}, translators~\cite{googletranslate, papago}, and paraphrasers~\cite{wordtune, quillbot}. 
However, NNESs face unique challenges when using AI writing assistants, particularly paraphrasing tools, \rev{which often generate multiple paraphrasing suggestions without explanations (Figure~\ref{fig:concept}).
This lack of explainability can be especially critical for NNESs who might lack the ability to independently assess the appropriateness of these suggestions.}

In this paper, we present findings from our preliminary interview study with 15 NNESs with \rev{varying levels of English proficiency and} experiences using AI writing assistants. Through our study, we highlight the specific challenges faced by NNE speakers when paraphrasing with AI writing assistants, and suggest potential design implications for enhancing the explainability of these tools. Although our study encompassed all AI writing assistants, our findings and suggestions are primarily concerned with the challenges associated with \rev{the paraphrasing tasks~(see Figure~\ref{fig:screenshot} and ~\ref{fig:screenshot:translator} in the Appendix for the illustration of paraphrasing capabilities).
For the rest of the paper, we refer to these AI writing assistants used in paraphrasing tasks as AI tools for simplicity.} 

\section{Findings from the interview}
\label{section:findings}

To understand the challenges faced by NNESs when using AI tools, we conducted a semi-structured interview with 15 NNESs who frequently write English emails and use AI tools. 
We first asked participants to choose one of the email scenarios used in the prior work~\cite{lim2022understanding} and compose an email with the writing tools they use on a daily basis in a think-aloud manner. We then asked participants about the challenges they faced while using the tools and their solutions to those challenges. To analyze the qualitative data gathered from the interviews, we transcribed all audio and screen recordings and conducted open coding and thematic analysis. See Appendix~\ref{appendix:interview} for a detailed description of the interview protocol and participants' backgrounds.

To improve their writing, many participants turned to the paraphrasing capabilities of AI tools and wished these tools could solve the challenges. Specifically, we observed that 12 participants used Grammarly~\cite{grammarly}, four used Wordtune~\cite{wordtune}, two used Quillbot~\cite{quillbot}, and one used Writefull~\cite{writefull} (see Appendix~\ref{appendix:screenshot:paraphrasing} for the illustration of the paraphrasing features of each tool). 
Seven participants also used translators such as Google Translate~\cite{googletranslate} and Naver Papago~\cite{papago} to paraphase the translated English outputs by perturbing their inputs in first language (see Appendix~\ref{appendix:screenshot:translate} for the illustration).


\subsection{Challenges Faced by NNESs in Writing} 
\label{section:findings:expectation}
The thematic analysis revealed that NNESs encounter unique challenges when writing in English, \rev{which often leads them to rely on AI tools to address these difficulties.} 
Out of 15 participants, 11 expressed difficulties in \textbf{accurately conveying their intention} in English sentences. 
Although NNESs knew the literal meaning of words, they often struggled with using those words in the appropriate context. For instance, P12 said: \textit{``In Korean, the words `ratio' and `proportion' are translated the same, so I wrote like, `The ratio of participants who are X is high.' My advisor got really angry at me and asked who uses `ratio' in this case.''}
Similarly, they had difficulties in \textbf{controlling the nuances and tones} of sentences, as P8 stated:\textit{``When it comes to emails or messages, the content is usually not too challenging. Nonetheless, I frequently find myself concerned about whether the nuances of my message align with my intended meaning. I worry that the message might come across as excessively polite or that it may be out of place with the given context.''} 
Five participants also commented that their \textbf{writing often sounds unnatural}. P11 said: \textit{``When I try to write something in English, I can put together a series of words that make sense, but it doesn't always sound natural. Instead, it may sound like the way sentences are structured in Korean, which I am not happy with.''}

\subsection{Challenges in Accepting Paraphrased Suggestions from AI Tools} 
\label{section:findings:challenge}

\rev{In our interview, NNESs had mixed feelings about the effectiveness of AI tools, noting that while these tools were helpful in addressing certain difficulties, they were not always effective in meeting their specific writing needs.}
The main challenge was that when presented with multiple paraphrased suggestions from AI tools, NNESs \textbf{could not confidently choose (or ``accept'') the best suggestion}.
In addition, it appeared that NNESs \textbf{did not fully trust AI tools}, as seven participants noted. \rev{Two participants, who were familiar with AI, were aware that the generative capabilities of AI are not guaranteed to be perfect, while the others learned through their experiences that AI tools often generated contextually inappropriate expressions as well as unnatural sentences with broken grammar and altered meanings.} Four participants were also cautious towards accepting \rev{suggestions} because they \textbf{did not know the rationale behind the suggestions provided by AI tools}. P6 questioned the reason for the changes in the paraphrased sentences: \textit{``I am not sure why Quillbot suggested changing the sentence like that.''} P8 also questioned what information the system is considering when generating suggestions: \textit{``I am suspicious whether the tool knows if I am writing an email right now and giving me suggestions that fit the context.''}
As a result, they viewed these tools as compromise solutions that were at least better than themselves. P9 noted: \textit{``When I use writing tools, I doubt the quality of my writing, but it is still better than writing without them. This is the best option for me.''}

\rev{NNESs, particularly those with lower levels of English proficiency, seemed to face greater difficulties in accepting paraphrased suggestions compared to those with high proficiency.}
Six participants mentioned that they found accepting the paraphrased suggestion challenging, especially because they \textbf{did not know the tones and nuances} of texts presented to them.
P4 noted: \textit{``Even if it is grammatically correct, I often feel uncertain whether this suggestion} is polite. For example, I am not sure if the question I have written in this email to my advisor sounds polite or rude.'' 
Moreover, they were \textbf{worried that the suggestion might have changed the meaning of their original text}. P8 noted, \textit{``I often doubt whether this suggestion is accurate. Take, for instance, the phrase `apply for leave' - while `leave' can refer to a vacation, I am concerned it could be interpreted in a different way, like quitting my job. I am worried that my boss might misunderstand my intentions.''}
In contrast, participants with higher levels of English proficiency reported less overhead in selecting the best suggestion. P10 stated: \textit{``I write about eight emails every day and read a bunch of messages from native speakers, so I know which expressions are commonly used and have no trouble picking the right suggestion.''}

\begin{figure*}[t]
    \centering
        \centering
        \begin{subfigure}[t]{0.229\linewidth}
            \centering
            \includegraphics[width=\linewidth]{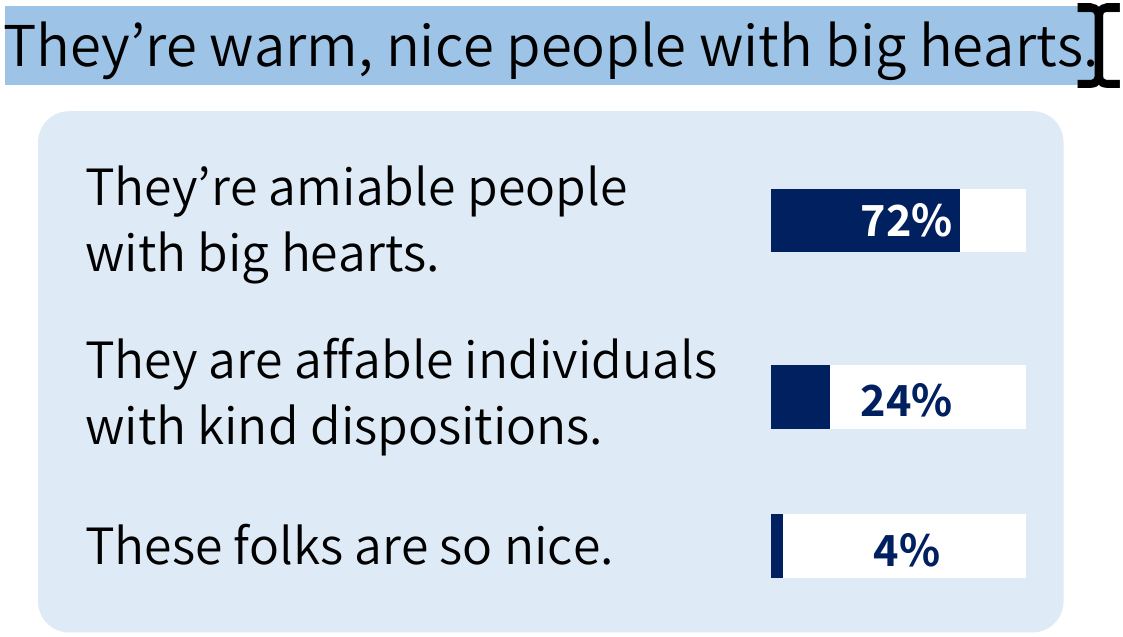}
            \caption{Model confidence}
            \label{fig:feature:raw}
        \end{subfigure}
        ~
        \begin{subfigure}[t]{0.27\linewidth}
            \centering
            \includegraphics[width=\linewidth]{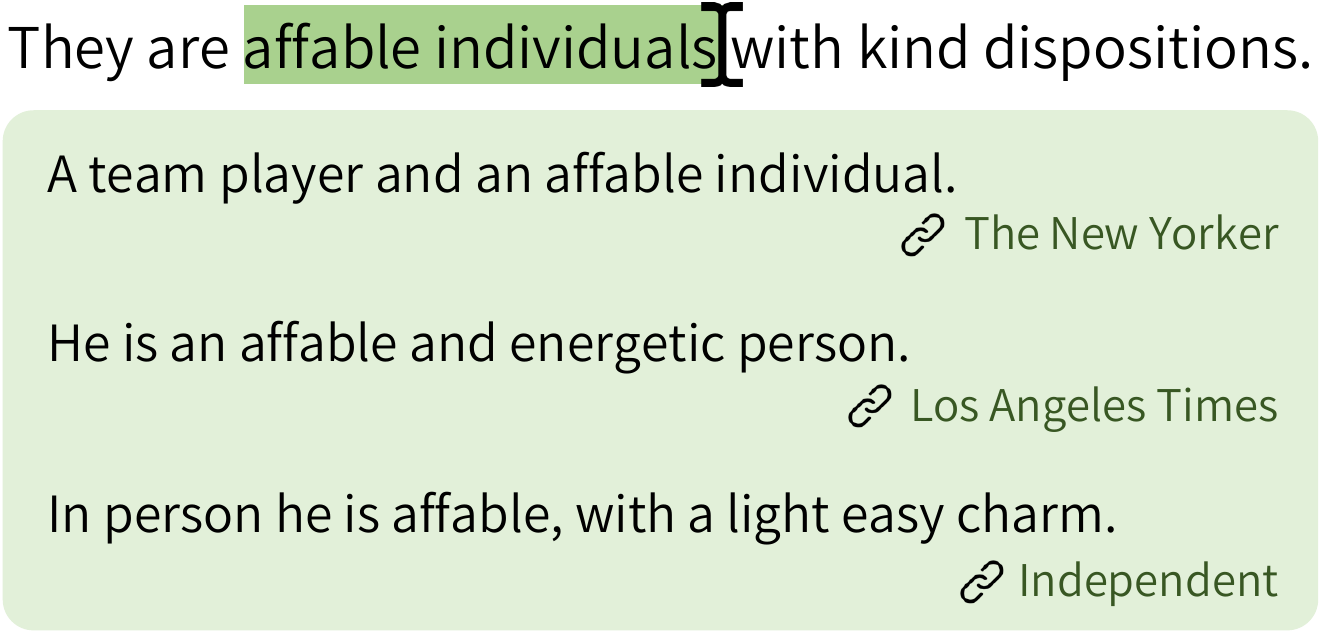}
            \caption{Examples from credible sources}
            \label{fig:feature:source}
        \end{subfigure}
        ~
        \begin{subfigure}[t]{0.21\linewidth}
            \includegraphics[width=\linewidth]{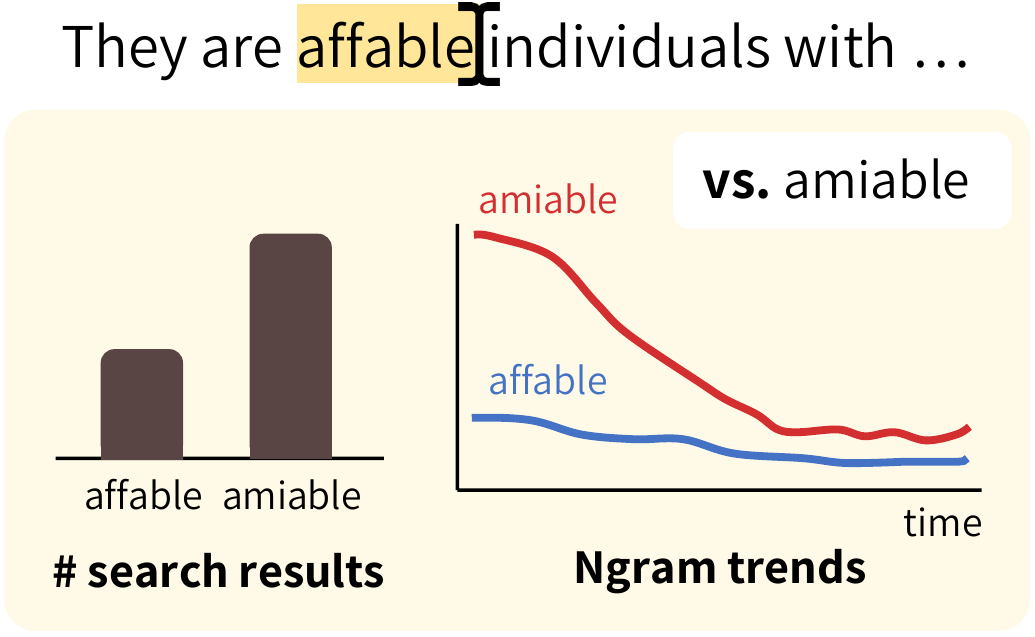}
            \caption{Data-driven usage trends}
            \label{fig:feature:stats}
        \end{subfigure}
        ~
        \begin{subfigure}[t]{0.252\linewidth}
            \includegraphics[width=\linewidth]{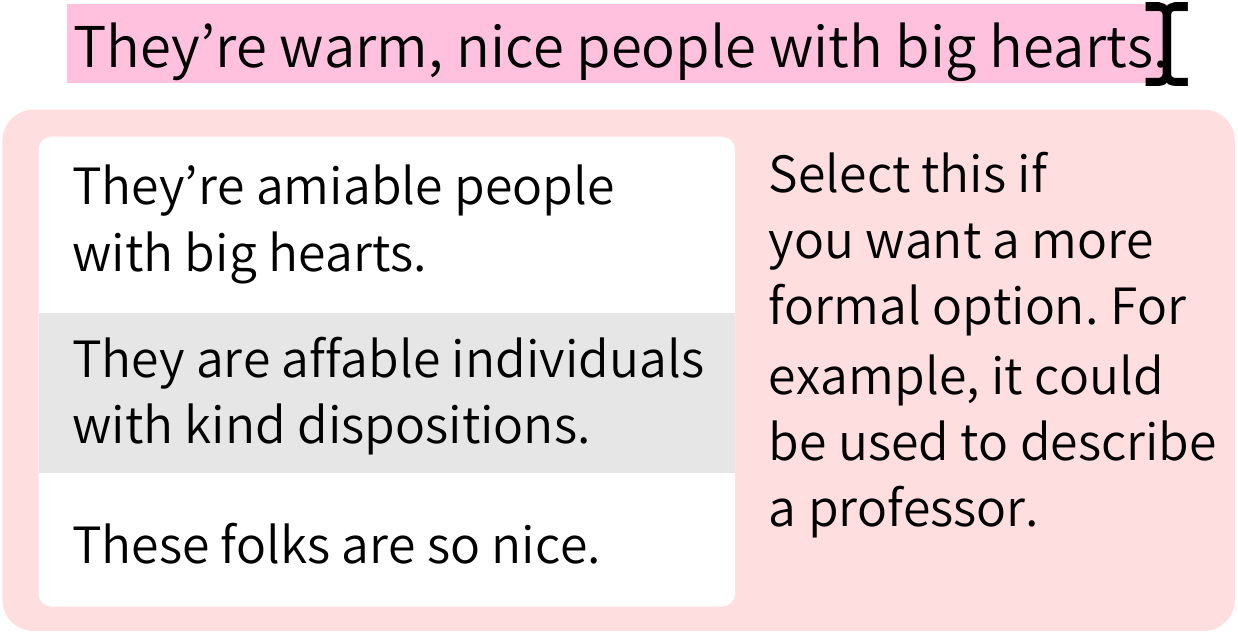}
            \caption{Textual explanations}
            \label{fig:feature:text}
        \end{subfigure}
    \vspace{-.2cm}
    \caption{Example user interfaces portraying design implications for non-native English speakers based on our findings. Explanations can be presented in the form of (a) confidence scores of model outputs, (b) example sentences from credible sources, (c) comparison of usage trends of original and paraphrased suggestions, and (d) textual explanation of the suggestions.}
    \vspace{-0.3cm}
    \label{fig:feature}
\end{figure*}

\vspace{-.1cm}
\subsection{NNESs' Strategies to Overcome Challenges} 
\label{section:findings:solution}
\looseness-1
When selecting the suggestion, participants used various strategies to aid their decision-making. The most common strategy used by eight participants was referring to \rev{\textbf{human-authored texts from credible sources}}. They referred to \rev{example sentences provided in} web sources such as dictionaries, Ludwig~\cite{ludwig}, and Thesaurus~\cite{thesaurus} to understand in what context and how the suggested words or expressions are used. P11 believed the example sentences are written by native English speakers: \textit{``I often refer to the Longman dictionary because it provides accurate information on how words are used in specific contexts. Its examples are taken from the native English corpus, so you can be confident that the words are used in the way that natives use them.''} 
Three participants, who are academic researchers, mentioned that they refer to relevant papers to see if particular phrases are frequently used in the literature. Two even found YouTube useful, as P5 stated, \textit{``I would search `I wish you are doing well' in YouTube to watch educational videos that explain the specific contexts this expression could be used. The fact that I can see the instructor's face, check their subscribers count, and read all the positive comments makes those videos much more trustworthy than a dictionary.''} 

Four participants appreciated \textbf{textual explanations} of suggestions, with three specifically noting Grammarly's explanation features, \rev{which not only suggest paraphrases but also explain why the suggested changes are necessary.} \rev{P2 stated: \textit{``When you read Grammarly's explanations, you can learn why you should make the corrections. It is helpful because it helps you become more careful when using similar expressions in the future.''}} P5, \rev{who did not use Grammarly,} wished for textual indicators that explain \textit{``the most appropriate contexts''} to use the expression as well as \textit{``the comparison between two expressions with similar meanings.''} P4 also wished to know the reasoning behind the suggestions from \rev{AI tools such as translators}, as he noted: \textit{``I wish the tool also generates the reason it is paraphrasing the original sentence like this.''}

Three participants found \textbf{statistical evidence} helpful. Specifically, two participants searched Google to refer to the number of search results. P11 explained:  \textit{``One way to find out if an expression is commonly used by native speakers is to search it on Google and see how many results you get.''} Moreover, P8 referred to Google's Ngram viewer~\cite{ngram} to see if the word or expression is trendy. 
Lastly, two participants mentioned model confidence scores. For instance, P9 appreciated Writefull~\cite{writefull} \rev{for displaying confidence scores alongside multiple suggestions, where the confidence score, represented as a percentage, indicates the likelihood that an AI's suggestion is accurate~\cite{10.1145/3351095.3372852} (see Figure~\ref{fig:screenshot:writefull} for an illustration of confidence scores).}
\rev{He stated: \textit{``If something has a really high confidence score, like 100\%, it is probably a good idea to pay attention to it and give it some extra consideration. But if it is more like 50\%, it is worth considering, while also keeping in mind that there might be other options you can explore.''}}
\rev{On the other hand,} P8 stopped using Wordtune~\cite{wordtune} because \textit{``there are so many suggestions, but I have no clue which one is the winner,''} which made him feel like \textit{``the suggestions are made carelessly,''} wishing for \textit{``a ranking system based on model confidence.''}

Seven participants mentioned they have given up on obtaining satisfactory results. P12 noted: \textit{``Grammarly does not fix everything, but that does not mean I am trying to improve the sentence any further. I just acknowledge that it might not be perfect, but it is the best I can do. Trying to fix all the errors on my own would cost me too much. Plus, getting the nuances of sentences right is something I mostly give up on. There are parts I do not know at all, so there is not much I can do about it.''} 
Furthermore, NNESs appeared to perceive the feedback provided by native English speakers to be the ultimate form of assistance. For important tasks such as paper writing or professional emails, three participants opted to seek human resources, such as native friends or proofreading services. 
P5 commented: \textit{``Whenever I ask my native friends for feedback, they usually say something like, `It is not wrong, but we usually say it like this.' So, when I am writing something important, I prefer to seek their help rather than relying on writing tools.''}

\section{Design Implications}
By drawing inspiration from explainable AI (XAI) literature~\cite{Gunning_Aha_2019} and language acquisition theory~\cite{krashen1981second}, we explore the design implications of AI tools and offer recommendations for addressing the challenges NNESs encounter identified in our interview.
Figure~\ref{fig:feature} presents four potential explanation features of AI tools.

\subsection{Revealing AI Models' Internal Mechanisms}
\label{section:discussion:transparency}
\looseness-1
\rev{Our study suggests that NNESs need additional support in accepting suggestions as they struggle to evaluate paraphrased suggestions of AI tools and question the rationale behind these suggestions.} 
A viable solution could involve elucidating the behavior of AI models by revealing their internal mechanisms. One such approach could involve presenting the \textbf{confidence scores} of model outputs (Figure~\ref{fig:feature:raw}), which indicate the probability of each model output and potentially provide users with a sense of when to trust or distrust the model~\cite{10.1145/3351095.3372852}. 
Previous research on prediction tasks has demonstrated that user performance improves with high accuracy indicators~\cite{10.1145/3287560.3287590} and user trust increases in high-confidence cases~\cite{10.1145/3351095.3372852}. \rev{However, caution should be exercised, as model confidence could be misleading~\cite{10.1145/3287560.3287590} and might result in low human trust in the model~\cite{10.1145/3290605.3300509}. For instance, high confidence score does not necessarily imply high accuracy and models can be overconfident in their predictions~\cite{10.1145/3290605.3300509}. Additionally, confidence scores might be calibrated using temperature scaling~\cite{pmlr-v70-guo17a}, which could affect the presentation of the confidence scores and accordingly the users' decisions.}

In addition to confidence scores, \rev{prior studies showed that visualizing \textbf{attention scores}~\cite{attention} of language models can help improve explainability~\cite{causal2017, exbert}.} Attention scores provide a measure of the importance of different input tokens when generating a specific output token. For example, visualizing the strength of dependencies between input-output tokens could help users understand which input tokens significantly impact the model's output, revealing the relationship between tokens and their influence on the generation~\cite{causal2017}. 
\rev{Such an approach could enhance NNESs' understanding of contextual associations in English by providing insights into the relationship between words in original and paraphrased texts. For example, a user might learn polite expressions by analyzing paraphrased words with high attention scores that correspond to polite expressions they are already familiar with.}

\subsection{Presenting Real-world Language Use Cases}
\label{section:discussion:implicit}

In this study, participants assessed the appropriateness of paraphrased suggestions by seeking real-world evidence, such as example sentences from credible sources or data-driven trends such as Google search results. This behavior aligns with \rev{the findings of} prior research that suggested creating a corpus of situation-specific facts is essential before generalizing across situations in learning tasks~\cite{solve1989chi}. Another line of research in the Natural Language Processing~(NLP) field, which aims to correct language learners' errors by retrieving example sentences~\cite{kaneko-etal-2022-interpretability} also complies with this finding. In line with this, we hypothesize that NNESs' tendency to assess suggestions using real-world examples reflects \textit{learning with understanding}. Specifically, NNESs engage in an \textit{implicit} language learning process~\cite{dekeyser2003implicit}, whereby individuals acquire language through exposure to \rev{language inputs and communicative interactions in real-life contexts.} 

As such, we propose that effective presentation of \textbf{example sentences} (Figure~\ref{fig:feature:source}) along with paraphrased suggestions could enhance NNESs' decision-making and language learning processes. 
\rev{A critical aspect to consider when selecting credible sources of sentences would be the use of corpora written by native English speakers. This consideration stems from the findings of our interview study, in which participants perceived texts written by native English speakers (e.g., Longman Corpus Network~\cite{longman}) as credible.}
Another interesting approach could be presenting use cases from \rev{\textbf{educational videos}}, as two participants found YouTube educational videos credible based on subscriber counts, positive comments, and the instructor's appearance. 
\rev{Supporting NNESs' decision-making could also be achieved by visualizing \textbf{data-driven usage trends} (Figure~\ref{fig:feature:stats}). Deriving quantitative evidence from big data could offer a straightforward approach to representing word prevalence. However, it is essential to carefully contextualize the information, as na\"ive search results might not be a reliable indicator of appropriateness. Various factors, including search algorithms and paid advertising, could influence these results~\cite{filterbubble}.}

\subsection{Generating Textual Explanations}
\label{section:discussion:explicit}

Integrating textual explanations (Figure~\ref{fig:feature:text}) into AI tools could foster natural and effective AI-powered support for NNESs. 
\rev{These explanations could take the form of local explanations to \textbf{provide the rationale for a single suggestion} and help users determine whether to trust a model on a case-by-case basis~\cite{lime}. Alternatively, the model could be given the context of writing, such as professional email writing or academic writing, and \textbf{generate feedback comments}~\cite{nagata-etal-2021-shared} on inappropriate user texts, explaining why certain expressions are unsuitable within the underlying context.
Another approach would involve \textbf{providing general explanations} that instruct users on the typical use cases of certain words or expressions. For example, when generating feedback comments on inappropriate usage of expressions, AI tools could also explain the contexts in which those expressions can be used. This approach can offer users an \textit{explicit} guidance~\cite{dekeyser2003implicit} on how language is used to convey meaning in different situations~\cite{Aufa_2016, ROSE2005385}. 
In addition, the effective presentation of textual explanations is a crucial consideration, particularly for NNESs who might struggle to comprehend the generated explanations due to limited English proficiency. Possible solutions include translating explanations into the user's first language or simplifying the language used in the explanations.}

\section{Discussion}
\label{section:discussion:considerations}

\subsection{Trade-off Between Efficiency and Quality}

\rev{In general, considering the trade-off between efficiency and quality is important in designing explanations~\cite{10.1145/3411764.3445372}. Providing \rev{complex} explanations (e.g., Figure~\ref{fig:feature:source}) may result in better suggestion selection and facilitate learning; however, it could take longer to select suggestions than simpler features (e.g., Figure~\ref{fig:feature:raw}). Conversely, while simple features may help users quickly select suggestions, they might be difficult to comprehend, resulting in lower quality and learning outcomes.
Moreover, users with varying levels of English proficiency and needs may perceive efficiency differently. Prior study~\cite{10.1145/3411764.3445372} have suggested that the cost of efficiency in accepting suggestions might be more burdensome for native speakers than NNESs. Similarly, we observed in our interview that participants with a higher level of English proficiency experienced \rev{less} overhead when selecting suggestions \rev{compared to those with low proficiency}. In such cases, providing explanations may not be necessary or may even hurt overall efficiency~\cite{10.1145/3411764.3445372}. }

\subsection{Learning Effects of Explanations}

\rev{Providing explanations to NNESs may facilitate their language learning~\cite{dekeyser2003implicit}; however, it is crucial to carefully design the explanations to maximize the learning effects of NNESs.
A key design consideration involves balancing explicit and implicit learning guidance to promote efficient language acquisition~\cite{doughty1998pedagogical}. Explicit learning guidance, such as textual instructions, could accelerate the learning process, but it may not always translate to flexible language application in diverse real-life contexts~\cite{dekeyser1998beyond, dekeyser2020skill}. On the other hand, implicit learning, such as providing example sentences, can be effective in helping NNESs generalize to new contexts~\cite{usage2017ellis} but may result in over-reliance on such examples~\cite{solve1989chi}. 
Additionally, an imbalanced amount and abstraction level of information could result in unsatisfactory learning experiences. Oversimplification of the explanations is known to increase the mental load on users and decrease their trust in the explanations~\cite{6645235}, as well as causing over-reliance on AI~\cite{vasconcelos2023explanations}. Conversely, presenting an excessive amount of information could overwhelm users~\cite{Roetzel2018InformationOI}. 
}

\subsection{Dialogue-based User Interaction}
\looseness-1
In addition to the conventional one-way interaction where human users accept or reject suggestions from AI tools, these assistants could also engage in a dialogue with users to better understand their needs and preferences. Recent advances in language models, such as ChatGPT \cite{chatgpt} and GPT-4 \cite{openai2023gpt4}, have already demonstrated the potential for dialogue-based user interaction. These models allow users to freely communicate with AI, providing them with greater control and flexibility to steer the model's output. However, this increased freedom also introduces new challenges, as NNESs might face challenges effectively communicating their intentions through English prompts, which could lead to suboptimal outputs or miscommunication between the user and AI. Future work can focus on assisting NNESs to effectively query the model. This may involve creating more NNE user-friendly interfaces or providing users with suggested prompts based on their specific needs. Additionally, research aimed at enhancing the robustness of AI models to variations in input can help mitigate the impact of poorly constructed prompts.

\subsection{Potential Bias AI Tools May Cause}
The design decisions involved in presenting explanations can unintentionally introduce further bias into AI tools. 
For instance, the manner in which an explanation is visualized or presented to the user could influence their decision-making process~\cite{beel2021unreasonable, cosley2003seeing}. This could lead to a biased text outcome, as recent work~\cite{jakesch2023co} suggests that collaborative writing with AI affects users' views, possibly affecting the contents of the written text. Moreover, NNESs might be particularly susceptible to this issue, as their lack of expertise or familiarity with the language might make them more vulnerable to the biases introduced by the AI outputs, leading to over-reliance~\cite{7349687}.

\begin{acks}
This work was supported in part by the Institute of Information \& Communications Technology Planning \& Evaluation (IITP) grant funded by the Korea government (MSIT) (No. 2022-0-00064, Development of Human Digital Twin Technologies for Prediction and Management of Emotion Workers’ Mental Health Risks).
\end{acks}

\bibliographystyle{ACM-Reference-Format}
\bibliography{05_references}

\newpage
\appendix

\section{Interview details}
\label{appendix:interview}

\subsection{Interview Protocol}
\label{appendix:interview:protocol}
We employed a semi-structured interview protocol to gather insights from participants regarding their email writing experiences. During the interview, the participants were first asked to select one of the provided email scenarios introduced in the prior work~\cite{lim2022understanding} (see Table~\ref{tab:emailscenarios} for the full list of scenarios) that they found most relevant to their personal or professional context. 
Participants were encouraged to write in their naturalistic settings, which may involve the use of AI writing assistants (e.g., Grammarly and Wordtune) or online searches (e.g., Thesaurus and dictionaries) to aid their writing process. Throughout the task, participants were asked to verbalize their thought processes in a think-aloud manner. Upon completion, we inquired the participants about their writing experiences, focusing on the challenges encountered during the session, strategies employed to overcome these difficulties, and suggestions for enhancing the writing tools they utilized. 

\begin{table*}[t]
\centering
\begin{tabular}{|l|l|}
\hline
1 &
  \begin{tabular}[c]{@{}l@{}}You ask your professor you took a class with a while ago to introduce you to someone who \\ may be hiring in your chosen career path.\end{tabular} \\ \hline
2 &
  \begin{tabular}[c]{@{}l@{}}You would like to take a week off from work to attend the wedding of a friend who lives \\ abroad. You are emailing your boss to ask for a week off.\end{tabular} \\ \hline
3 &
  \begin{tabular}[c]{@{}l@{}}You have an interview for a new job next week, but you realize you cannot make it on the \\ scheduled date. You are emailing the hiring manager to reschedule.\end{tabular} \\ \hline
4 &
  \begin{tabular}[c]{@{}l@{}}You are interested in working at a particular company and heard that an alumnus of your \\ college currently works there. You are emailing this person to ask if they would speak with \\ you about their company in a video call.\end{tabular} \\ \hline
5 &
  \begin{tabular}[c]{@{}l@{}}You are in charge of a fundraising campaign at work to help needy children. You are \\ emailing your colleagues to ask them to donate to this campaign.\end{tabular} \\ \hline
6 &
  \begin{tabular}[c]{@{}l@{}}You received an email from your apartment management office fining you for not recycling \\ properly. You believe the fine is an error and are emailing the office to ask them to cancel it.\end{tabular} \\ \hline
\end{tabular}
\vspace{1em}
\caption{The six email scenarios we used in the interview study. We borrowed the scenarios from the previous literature~\cite{lim2022understanding}.}
\label{tab:emailscenarios}
\end{table*}

\vspace{-.2em}

\subsection{Participant Details}
\label{appendix:interview:participants}

We recruited 15 NNE speakers (five females, four males, six preferred not to say) who engage in English writing activities at least once a week. The participants included six undergraduate students, seven graduate students, one research assistant, and one information technologist. Participants were recruited through social networking services. Each participant received a compensation of 35,000 KRW (approximately 27 USD) for participating in the study. The study was approved by the Institutional Review Board of the first author's institution. Table~\ref{tab:participants} shows the detailed background of the interview participants we recruited. When recruiting participants, we asked the participants to self-assess their English proficiency according to the Common European Framework of Reference for Languages (CEFR)~\cite{council2001common} measurement. CEFR levels are defined as basic (beginner and elementary levels; A1 and A2), independent (intermediate and upper intermediate levels; B1 and B2), and proficient (advanced and proficiency levels; C1 and C2). We provided the following rubrics to the participants for rating one's English proficiency. As a result, we recruited two basic-level, seven independent-level, and six proficient-level participants. 

\begin{itemize}
    \item A1 (Beginner): You can understand and use basic phrases and expressions. You can communicate in simple ways when people speak slowly to you.
    \item A2 (Elementary): You can take part in simple exchanges in familiar topics. You can understand and communicate routine information.
    \item B1 (Intermediate): You can communicate in situations and use simple language to communicate feeling, opinions, plans and experiences.
    \item B2 (Upper Intermediate): You can communicate easily with native English speakers. You can understand and express some complex ideas and topics.
    \item C1 (Advanced): You can understand and use a wide range of language. You can use English flexibly and effectively for social and academic purposes.
    \item C2 (Proficiency): You can understand almost everything you hear or read. You can communicate very fluently and precisely in complex situations.
\end{itemize}

\begin{table*}[t]
\centering
\renewcommand{\arraystretch}{1.1}
\begin{tabular}[t]{lcllll} 
\toprule
 & \begin{tabular}[c]{@{}l@{}}English \\ Proficiency\end{tabular} & \begin{tabular}[c]{@{}l@{}}First \\ Language\end{tabular} & Residence & AI Writing Assistants & \begin{tabular}[c]{@{}l@{}}Profession \\ (Major)\end{tabular} \\ 
 \midrule
P1 & A1 & Korean & South Korea & Grammarly, Naver Papago & \begin{tabular}[c]{@{}l@{}}Undergraduate student \\ (Electrical Engineering)\end{tabular} \\
\rowcolor[HTML]{EFEFEF} 
P2 & A2 & Korean & South Korea & Grammarly, Naver Papago & \begin{tabular}[c]{@{}l@{}}Undergraduate student \\ (Computer Science)\end{tabular} \\
P3 & B1 & Korean & South Korea & Naver Papago, Google Docs AutoCorrect & \begin{tabular}[c]{@{}l@{}}Undergraduate student \\ (Medicine)\end{tabular} \\
\rowcolor[HTML]{EFEFEF} 
P4 & B1 & Korean & United States & \begin{tabular}[c]{@{}l@{}}Grammarly, Google Translate, Naver Papago, \\ Google SmartCompose, Tabnine\end{tabular} & \begin{tabular}[c]{@{}l@{}}Graduate student \\ (Information Science)\end{tabular} \\
P5 & B1 & Korean & Sweden & \begin{tabular}[c]{@{}l@{}}Google Translate, Google Docs AutoCorrect,\\ Google SmartCompose\end{tabular} & \begin{tabular}[c]{@{}l@{}}Undergraduate student \\ (Education)\end{tabular} \\
\rowcolor[HTML]{EFEFEF} 
P6 & B2 & Chinese & United States & Grammarly, Google Translate, Quillbot & \begin{tabular}[c]{@{}l@{}}Research assistant \\ (Human-Computer Interaction)\end{tabular} \\
P7 & B2 & French & United States & Grammarly, Google Translate, Wordtune & Information technologist \\
\rowcolor[HTML]{EFEFEF} 
P8 & B2 & Korean & South Korea & \begin{tabular}[c]{@{}l@{}}Grammarly, Google Translate, Naver Papago, \\ Kakao i Translate, Wordtune\end{tabular} & \begin{tabular}[c]{@{}l@{}}Graduate student \\ (Human-Computer Interaction)\end{tabular} \\
P9 & B2 & Korean & South Korea & \begin{tabular}[c]{@{}l@{}}Grammarly, Google Translate, Naver Papago, \\ Kakao i Translate, Google SmartCompose,\\ Writefull\end{tabular} & \begin{tabular}[c]{@{}l@{}}Graduate student \\ (Computer Science)\end{tabular} \\
\rowcolor[HTML]{EFEFEF} 
P10 & C1 & Korean & United States & Grammarly, Wordtune, Google SmartCompose & \begin{tabular}[c]{@{}l@{}}Graduate student \\ (Social Science)\end{tabular} \\
P11 & C1 & Korean & United Kingdom & Wordtune & \begin{tabular}[c]{@{}l@{}}Graduate student \\ (Psychology)\end{tabular} \\
\rowcolor[HTML]{EFEFEF} 
P12 & C1 & Korean & South Korea & \begin{tabular}[c]{@{}l@{}}Grammarly, Google Translate, \\ Google SmartCompose\end{tabular} & \begin{tabular}[c]{@{}l@{}}Graduate student \\ (Computer Science)\end{tabular} \\
P13 & C2 & Korean & South Korea & Grammarly & \begin{tabular}[c]{@{}l@{}}Undergraduate student \\ (Business)\end{tabular} \\
\rowcolor[HTML]{EFEFEF} 
P14 & C2 & Filipino & South Korea & Grammarly, Google Translate, Naver Papago & \begin{tabular}[c]{@{}l@{}}Graduate student \\ (Electrical Engineering)\end{tabular} \\
P15 & C2 & Hindi & India & Grammarly, Quillbot, LanguageTool & \begin{tabular}[c]{@{}l@{}}Undergraduate student \\ (Information Technology)\end{tabular} \\ \bottomrule
\end{tabular}
\vspace{1em}
\caption{Detailed background information of the interview participants. The participants used various sets of AI writing assistants, which include Grammarly~\cite{grammarly}, Google Translate~\cite{googletranslate}, Naver Papago~\cite{papago}, Google SmartCompose~\cite{smartcompose}, Wordtune~\cite{wordtune}, Google Docs AutoCorrect~\cite{gdocautocorrect}, Kakao i Translate~\cite{kakaoi}, Quillbot~\cite{quillbot},  Writefull~\cite{writefull}, LanguageTool~\cite{languagetool}, and Tabnine~\cite{tabnine}. Refer to Section~\ref{appendix:interview:participants} in Appendix for the descriptions of english proficiency levels.}
\label{tab:participants}
\end{table*}

\section{AI Writing Assistants Details}
\label{appendix:screenshot}

\subsection{Paraphrasing Functionalities}
\label{appendix:screenshot:paraphrasing}
\begin{figure*}[t]
\captionsetup[subfigure]{justification=centering} 
    \centering
    \begin{subfigure}[t]{1\textwidth}
        \centering
        \begin{subfigure}[t]{0.58\textwidth}
            \centering
            \includegraphics[width=0.95\linewidth]{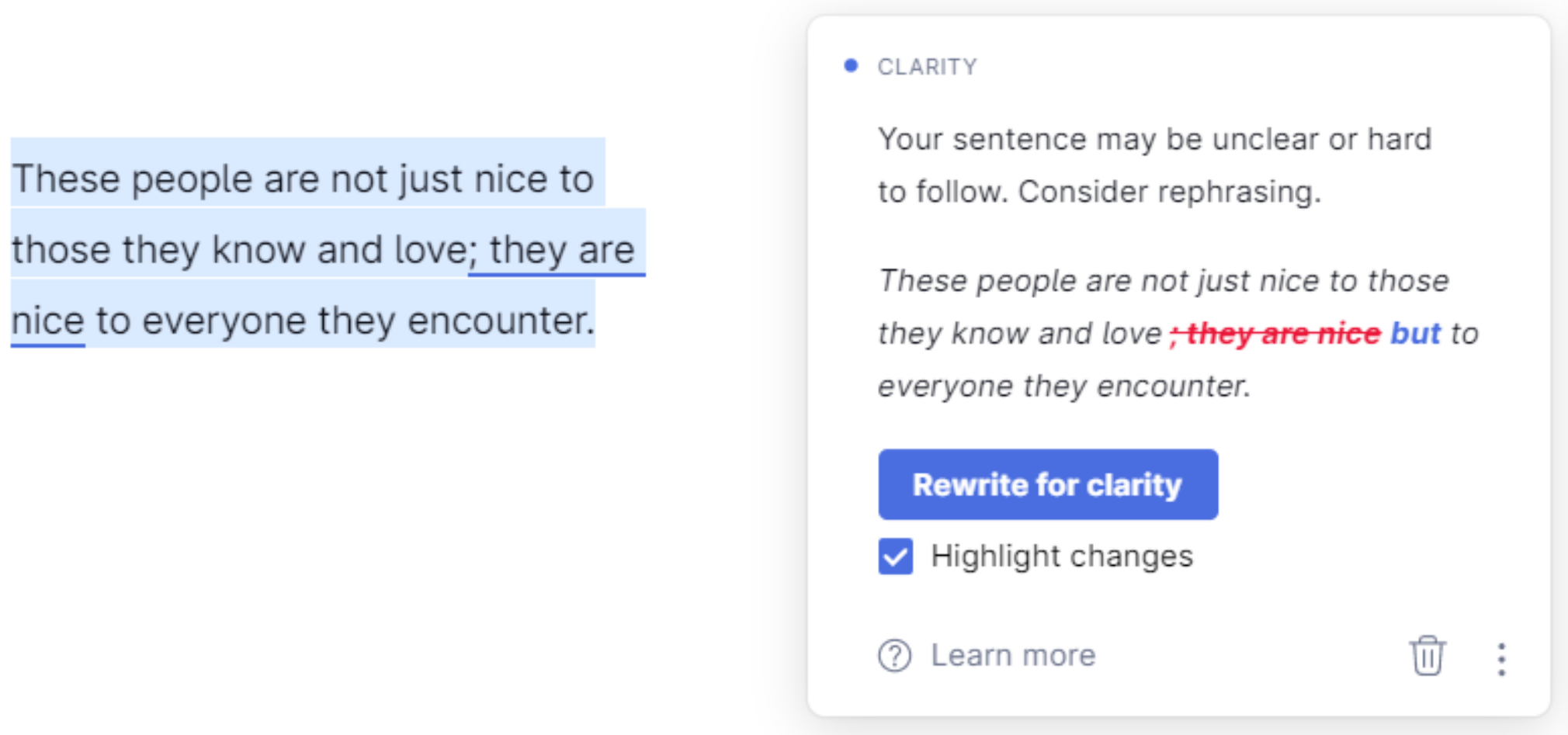}
            \caption{Grammarly~\cite{grammarly}}
            \label{fig:screenshot:grammarly}
        \end{subfigure}
        ~
        \begin{subfigure}[t]{0.42\textwidth}
            \centering
            \includegraphics[width=0.95\linewidth]{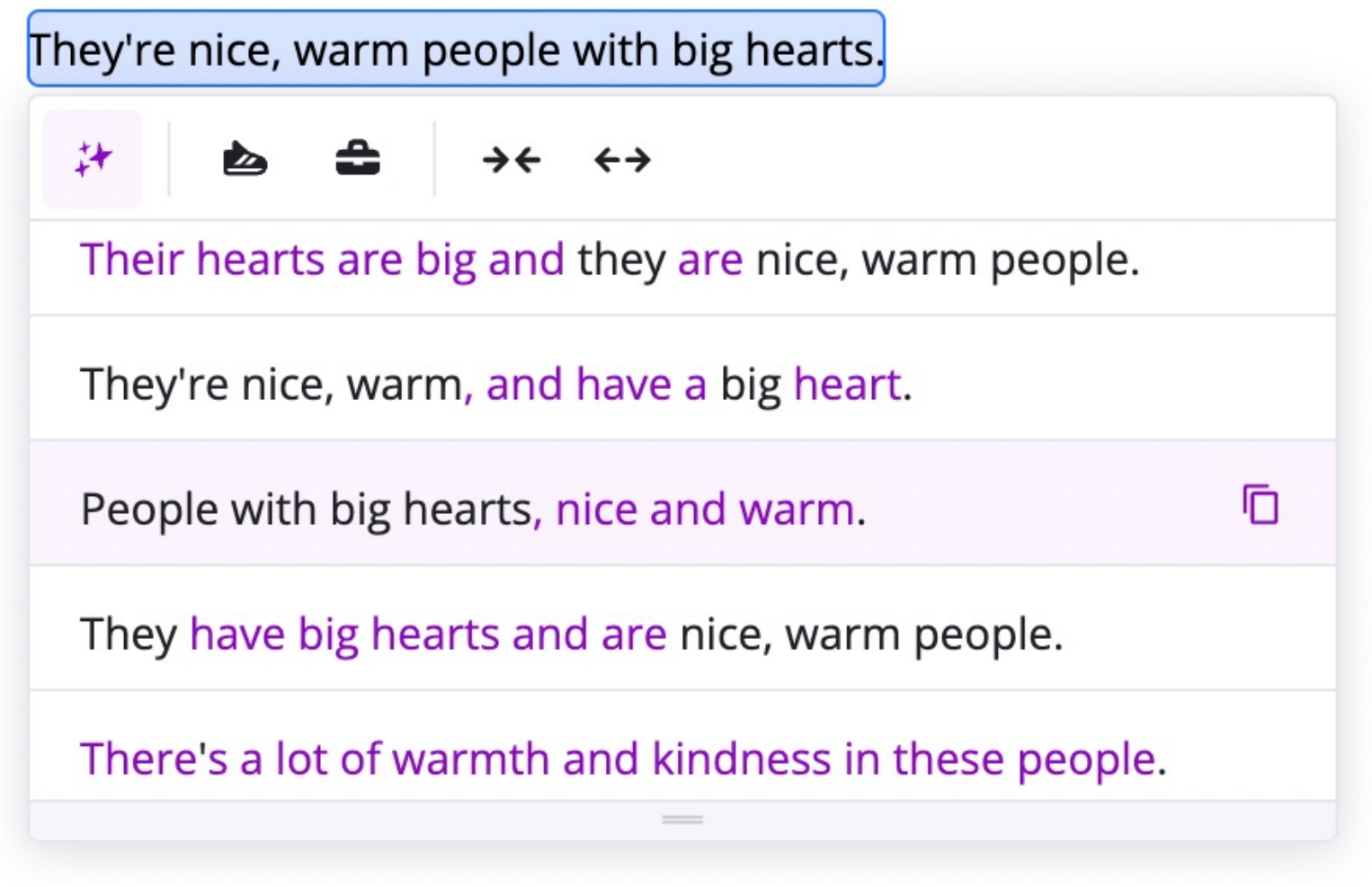}
            \caption{Wordtune~\cite{wordtune}}
            \label{fig:screenshot:wordtune}
        \end{subfigure}
    \centering
    \end{subfigure}

    \vspace{1em}
    
    \begin{subfigure}[t]{1\textwidth}
        \centering
        \begin{subfigure}[t]{0.58\textwidth}
            \centering
            \includegraphics[width=0.95\linewidth]{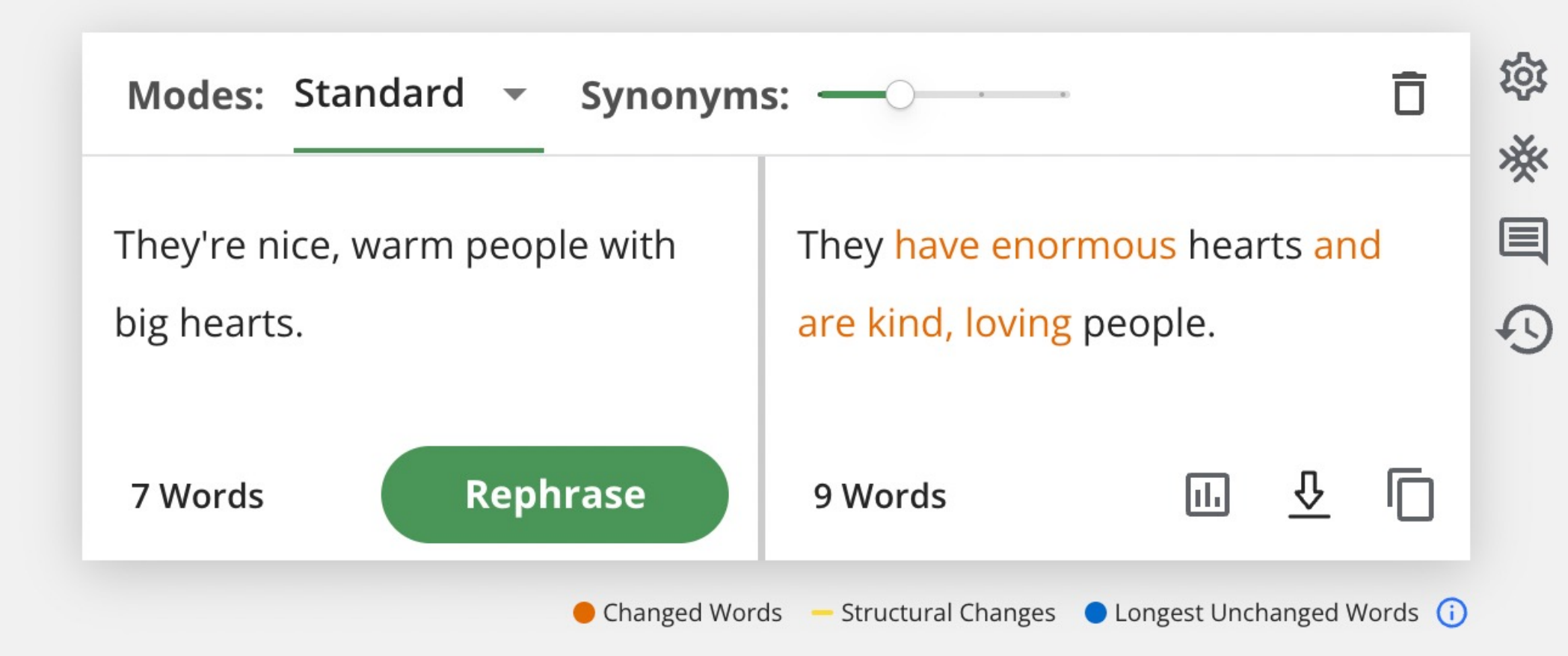}
            \caption{Quillbot~\cite{quillbot}}
            \label{fig:screenshot:quillbot}
        \end{subfigure}
        ~
        \begin{subfigure}[t]{0.42\textwidth}
            \centering
            \includegraphics[width=0.95\linewidth]{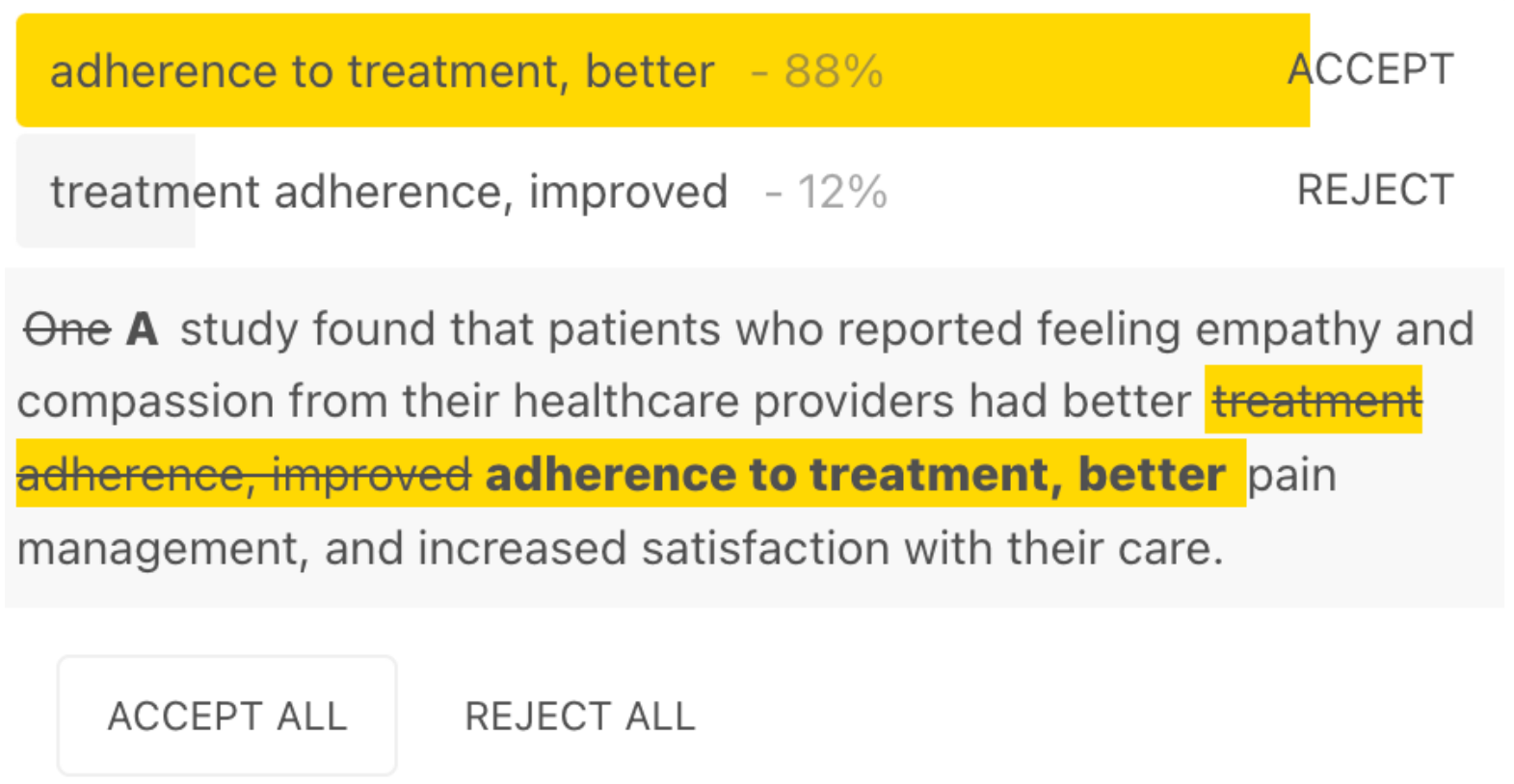}
            \caption{Writefull~\cite{writefull}}
            \label{fig:screenshot:writefull}
        \end{subfigure}
    \centering
    \end{subfigure}
    
    \vspace{1em}
    \caption{Screenshot of the paraphrasing features provided in each AI writing tools. (a) When a user clicks underlined words or phrases, Grammarly~\cite{grammarly} displays an explanation of the issue and offers suggestions for improvement. Users can either accept the suggestion by clicking ``Rewrite for clarify'' button or reject it by clicking the trashcan icon on the bottom right corner of the display. (b) Wordtune~\cite{wordtune} suggests multiple paraphrased suggestions for each input. Users can accept or copy one of the suggestions, or dismiss the suggestions. (c) Quillbot~\cite{quillbot} suggests a single paraphrased suggestion, where users can accept it or regenerate the suggestion by clicking ``Rephrase'' button. (d) When a user clicks underlined words or phrases, Writefull~\cite{writefull} shows a paraphrased suggestion and compares the original and paraphrased versions by showing confidence scores in percentage.}
    \label{fig:screenshot}
\end{figure*}

Figure~\ref{fig:screenshot} provides the screenshots of the paraphrasing features in the AI writing tools (Grammarly~\cite{grammarly}, Wordtune~\cite{wordtune}, Quillbot~\cite{quillbot}, and Writefull~\cite{writefull}) that the participants mentioned during the interview. 

\subsection{Using Translators as Paraphrasers}
\label{appendix:screenshot:translate}
\begin{figure*}[t]
    \centering
        \centering
        \begin{subfigure}[t]{0.45\linewidth}
            \centering
            \includegraphics[width=0.9\linewidth]{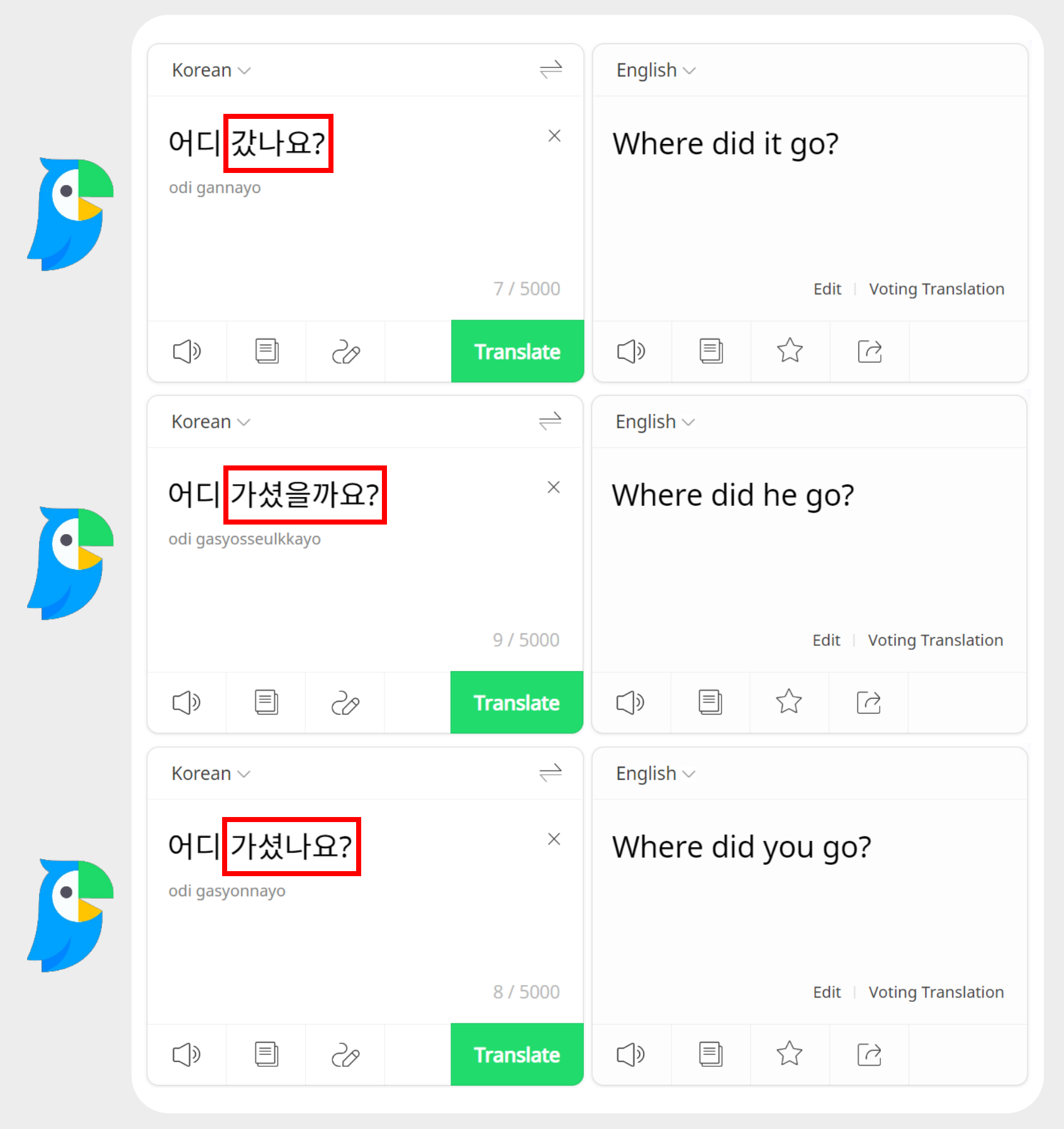}
            \caption{Perturbing L1 inputs.}
            \label{fig:screenshot:translator:perturb}
        \end{subfigure}
        ~
        \begin{subfigure}[t]{0.45\linewidth}
            \centering
            \includegraphics[width=0.965\linewidth]{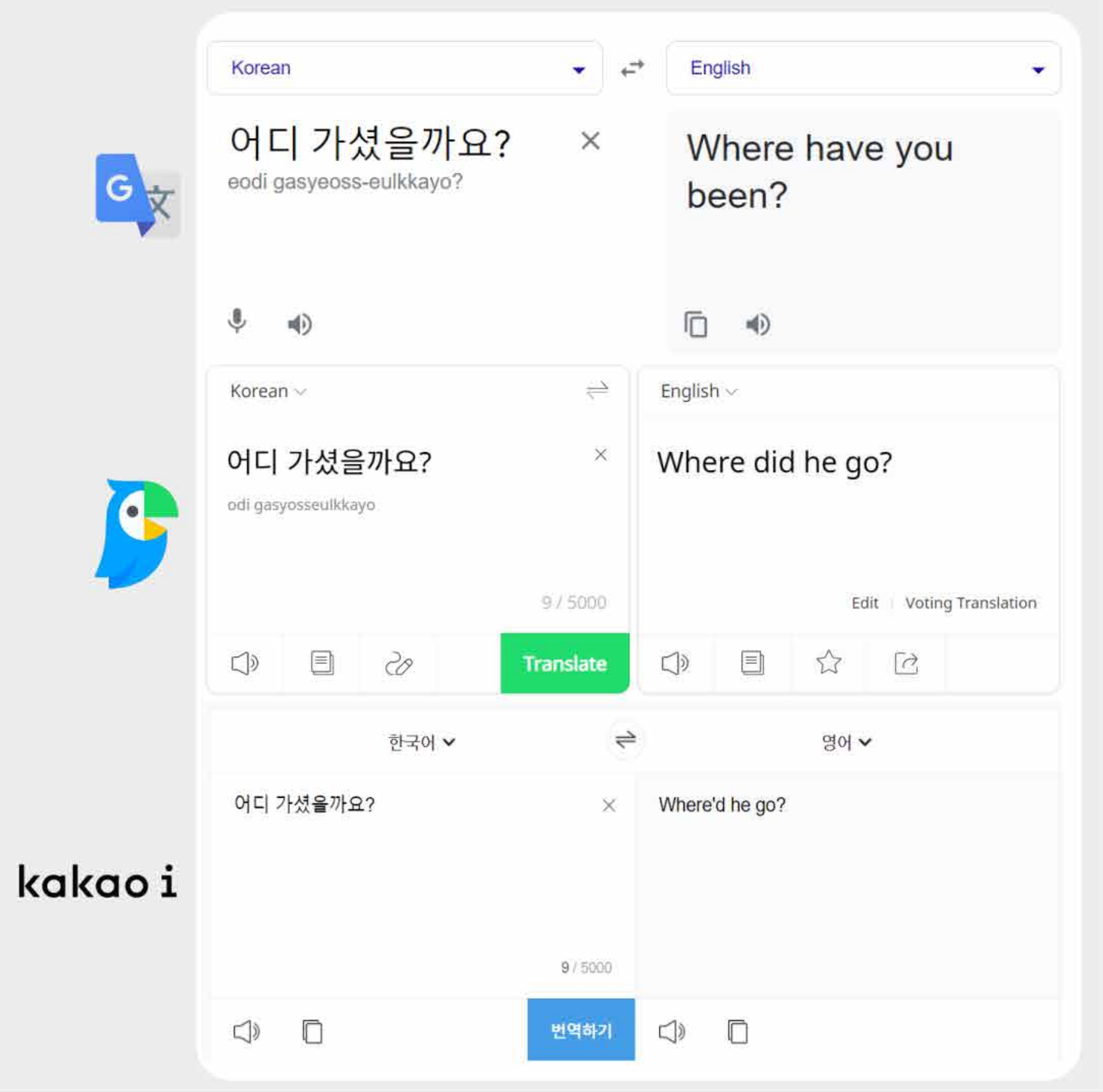}
            \caption{Using multiple translators.}
            \label{fig:screenshot:translator:multiple}
        \end{subfigure}
    \vspace{-.2cm}
    \caption{Illustration of how interview participants used translators as paraphrasing tools. In (a), only the honorific forms of texts in first language (L1; in this example, Korean) are perturbed (perturbations are indicated by red boxes) and translated using Naver Papago~\cite{papago}. In (b), the same L1 input was translated using Google Translate~\cite{googletranslate}, Naver Papago~\cite{papago}, and Kakao i Translate~\cite{kakaoi}.}
    \vspace{-0.3cm}
    \label{fig:screenshot:translator}
\end{figure*}

Figure~\ref{fig:screenshot:translator} illustrates how users used translators as paraphrasing tools. Users employed two strategies to use translators as paraphrasing tools, either by perturbing their inputs in first language within a single translator or by translating the same L1 input with different systems. In our interview, participants used among Google Translate~\cite{googletranslate}, Naver Papago~\cite{papago}, and Kakao i Translate~\cite{kakaoi}. 

\end{document}